%% file: main.tex
\documentclass{article} % For LaTeX2e

\usepackage[square,numbers]{natbib}
\usepackage[nonatbib,preprint]{neurips_2024}
\usepackage{graphicx} 
\usepackage{microtype}

\usepackage{url}
\usepackage{booktabs}
\usepackage{xspace}
\usepackage{bbm}
\usepackage[table]{xcolor}
\usepackage{colortbl}
\usepackage{comment}
    
\usepackage{wrapfig}
\usepackage{float}
\usepackage{amsmath}

\definecolor{columbiablue}{rgb}{0.61, 0.87, 1.0}

\usepackage{soul}
\usepackage{hyperref}  % hyperref package can sometimes cause issues with linking citations to the bibliography. Make sure you it's last package in preamble

\title{ColorGrid: A Multi-Agent Non-Stationary Environment for Goal Inference and Assistance}

\author{
    Andrey Risukhin$^{\diamond}$\thanks{Equal contribution.} \enspace 
    Kavel Rao$^{\diamond*}$ \enspace 
    Ben Caffee$^{\diamond}$\thanks{Equal contribution.} \enspace 
    Alan Fan$^{\diamond\dagger}$ \enspace \\
    $^{\diamond}$University of Washington \\ 
    \texttt{\{risuka, kavelrao, alanfan, bncaffee\}@cs.washington.edu}
}

\newcommand{\envname}{\textsc{ColorGrid}\xspace}

\begin{document}

\maketitle

\input{sections/00_abstract.tex}
\input{sections/01_introduction.tex}
\input{sections/02_related_work.tex}
\input{sections/03_method.tex}
\input{sections/04_experiments.tex}
\input{sections/05_discussion}
\input{sections/06_acknowledgements.tex}

\newpage

\bibliography{main}

\newpage

\appendix
\input{sections/appendix.tex}

\end{document}

%% file: sections/00_abstract.tex
\begin{abstract}

Autonomous agents' interactions with humans are increasingly focused on adapting to their changing preferences in order to improve assistance in real-world tasks. Effective agents must learn to accurately infer human goals, which are often hidden, to collaborate well. However, existing Multi-Agent Reinforcement Learning (MARL) environments lack the necessary attributes required to rigorously evaluate these agents' learning capabilities. To this end, we introduce \envname, a novel MARL environment with customizable non-stationarity, asymmetry, and reward structure. We investigate the performance of Independent Proximal Policy Optimization (IPPO), a state-of-the-art (SOTA) MARL algorithm, in \envname and find through extensive ablations that, particularly with simultaneous non-stationary and asymmetric goals between a ``leader'' agent representing a human and a ``follower'' assistant agent, \envname is unsolved by IPPO. To support benchmarking future MARL algorithms, we release our environment code, model checkpoints, and trajectory visualizations at \href{https://github.com/andreyrisukhin/ColorGrid}{https://github.com/andreyrisukhin/ColorGrid}.

\end{abstract}

%% file: sections/01_introduction.tex
\section{Introduction}

The integration of autonomous assistants into real-world tasks with humans necessitates robust guidance and correction. Nascent work is examining how to train autonomous agents to collaborate with humans of diverse skill levels, yet these studies always assumed a fixed objective. 
In the real world, human goals change -- consider a surgeon encountering unexpected complications during a surgery. A robotic surgery assistant must seamlessly follow suit, adapting on-the-fly.
As such, we contend that training collaborative agents to adaptively help humans in real-world tasks requires adapting to the human’s hidden and changing goals. 
We develop a benchmark that tests this aspect of human-AI coordination, creating a test-bed that enables focusing on the core problems of inferring goals in real-time. Agents that can make progress on this benchmark take a step toward helping humans in the real-world.

Many prior works explore zero-shot coordination with other agents and with humans \citep{dennis2021emergent, strouse2022collaborating, jeon2020shared, Nikolaidis2017HumanrobotMA}, but their goals are typically modeled as static throughout a given episode. 
Other works study social agents and their collaboration when augmented with message passing \cite{weil2024generalizability, Shrestha_2014}, but in reality we seek autonomous agents that do not require explicit messages to seamlessly and quickly transition between goals when assisting humans with arbitrary tasks. In settings such as elder care, it is unrealistic to expect humans to describe new goals in detail to the assistant, since the tasks may be time sensitive and require immediate attention.

\begin{figure}[h]
\begin{center}
\caption{\envname visualization demonstrating how assistant agent (follower) learns after observing human (leader) actions.}
\includegraphics[width=\linewidth]{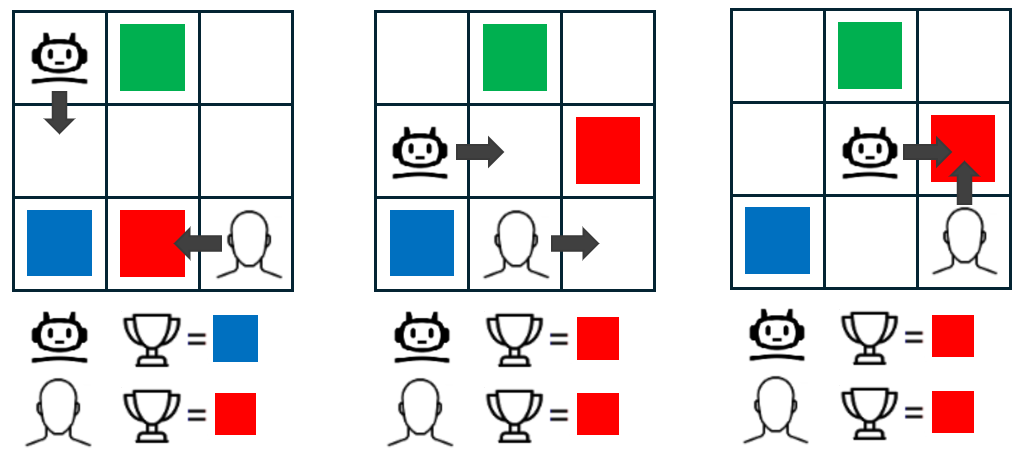}
\end{center}
    \label{fig:environment}
\end{figure}

In competitive Multi-Agent Reinforcement Learning (MARL) domains, self-play \cite{silver2017mastering} has been shown to reach human-level play without human data using model-free reinforcement learning. Self-play excels in competition, but works poorly for \emph{cooperation}. It produces agents that can only coordinate with themselves, and thus break or display ``stubbornness'' when needing to work with other agents who have different behavior. 
In contrast, population play \cite{Jaderberg2018HumanlevelPI} trains a group of agents who play in teams or against each other; however, works using these methods typically treat the high-level goal as static throughout an episode and focus on coordination at the strategy level. 

In this work, we present our two-fold contribution to address the lack of focus on learning to infer non-stationary goals: (1) Releasing \envname, a novel MARL environment with configurable levels of non-stationarity, reward sparsity, asymmetry, and reward structure and (2) Exploring how Independent Proximal Policy Optimization (IPPO) \cite{schulman2017proximal}, a SOTA MARL algorithm \cite{dewitt2020independentlearningneedstarcraft}, performs in \envname with broad ablations on environment configurations, reward shaping, and learning parameters. We demonstrate that with non-stationary goals and without providing explicit goal information to the ``follower'' autonomous assistant, \envname is a challenging MARL environment currently unsolved by IPPO which has been shown to perform well in collaborative settings \cite{yu2021surprising}. Particularly, we find that IPPO is insufficient even for the baseline when the follower is explicitly given the goal and the cost to explore is set too high.

The inability of SOTA MARL algorithms to address the problems of coordinating with a partner in a non-stationary environment with hidden goals highlight the fact that additional research is needed into MARL algorithms that can truly assist human partners. We hope that, by sharing the code for this benchmark, we can spur more research progress towards real-world human-AI cooperation.

%% file: sections/02_related_work.tex
\section{Related Work}

\paragraph{Zero-shot coordination with humans \& Ad-hoc Team-play}
Multiple approaches exist for zero-shot coordination with humans and ad-hoc team-play.
To encourage strategic diversity while still averting the need to collect human data, fictitious co-play \cite{strouse2022collaborating} trains agents against a population of self-play agents and prior training checkpoints in the game Overcooked \cite{overcooked}. Several works in robotics have also modeled shifting human goals to improve robot-human cooperation. For instance, \citet{jeon2020shared} develop a shared autonomy method for precise robotic manipulation, reinterpreting user inputs in a latent space based on a confidence-conditioned prediction of the human's goal. In the setting of human-robot cooperation where the two agents may prefer different strategies to achieve a given task, \citet{Nikolaidis2017HumanrobotMA} model the human's adaptability as a latent variable in a POMDP (Partially Observable Markov Decision Process) \cite{Smallwood1973TheOC} and adjust the robot's weighting between its and the human's preferred strategies to maximize the team's rewards depending on the human's capability to switch modes. Shared autonomy, a robot estimating operator intent based on operator inputs, can benefit from modeling those intents as Bayesian \cite{recursivebayesian}. Moreover, prior work such as Yell At Your Robot \cite{shi2024yell} explore real-time corrections of low-level robot actions through natural language. While providing impressive adaptability, this approach requires a human to constantly monitor the robot for potential mistakes, which is unrealistic in some downstream applications. Most of these works model human intent explicitly with hand-crafted heuristics. Our contribution is a step towards extending this adaptability to implicit higher-level goals in truly cooperative environments where both humans and autonomous agents collaborate without the need for explicit feedback. 

\paragraph{Multi-Agent Environments}

Distinct multi-agent environments have been designed to understand the challenges of agent interactions. This section outlines the main benefits of \envname over other environments. The popular Overcooked environment \cite{overcooked} is where agents (autonomous and human) use shared ingredients to produce and deliver soup. Across different kitchen layouts, collaboration between agents is explicitly required to complete the task, including both high-level strategic and low-level mobility collaboration. In contrast, \envname involves the complexity of having a dynamic objective via goal block switching, within an episode, rather than agents coordinating how to share resources, which could potentially be more complex. The Goal Cycle environment built by \citet{ndousse2021emergent} is a 13x13 grid containing goal and obstacle tiles, which are randomized at the start of each episode; agents receive rewards for traversing the goal tiles in a specific \textit{hidden} order and receive penalties for following the incorrect order. While the goal order is hidden, the environment is stationary, unlike \envname.

The Multi-Agent Particle Environments (MPE) introduced by \citet{mordatch2018emergence} initialize every agent with one or more private, physically-grounded goals that may require cooperation from other agents. Examples of goals in this environment include going to a landmark and looking at a location. Additionally, there is a channel through which agents can broadcast messages to others for a small cost. MPE use a continuous representation of physical locations while representing time as discrete. Distinct from MPE's vector world, level-based foraging environments (LBF) \cite{albrecht2015gametheoretic} are represented in a discrete grid-world. Our mechanism for assigning penalty fundamentally differs from level-based foraging environments, which only give positive rewards from collecting items but impose an additional penalty every step to incentivize rapid collection. 

The Multi-Robot Warehouse environment \cite{epymarl} (MRW) challenges agents to learn long chains of actions to complete deliveries and earn sparse rewards. While these environments have implicit non-stationarity from containing multiple agents, \envname introduces non-stationarity from the actual goal changing. \envname also doesn't have a communication channel between agents, i.e. the "follower" agent isn't notified when the "leader" "learns" of the goal switching since in real life scenarios it's not always possible to communicate, due to inaccessibility for instance.

The Coingrid environment \cite{raileanu2018modeling} evaluates two symmetric agents on a fully cooperative task. Specifically, there is an 8$\times$8 grid containing 4 coins for each of 3 colors. The two agents are each randomly assigned a color, with the goal of collecting either its assigned color or its partner agent's color to collectively maximize the reward without picking up non-assigned colors. While Coingrid does have hidden goals, our approach focuses on the case where those goals can change (and can also be hidden).

The Watch-And-Help (WAH) challenge \cite{puig2021watchandhelp} tests how much an agent Bob can accelerate task completion after observing another agent, Alice, demonstrate the task end-to-end. WAH explicitly enumerates the goals of a task as a set of predicates and their counts. In contrast, our motivation is to solve a continuous task without an explicit description. Rather than clearly delineating the start and end of a task, we design \envname to encourage a helper agent to act as soon as the intended goal is clear. Furthermore, observing a task to infer the goal and then helping a human finish the same task as fast as possible in a different episode is not an evaluation that supports our goal of real-time, continuous adaptation within an episode.

%% file: sections/03_method.tex
\section{Method}
\subsection{Environment}

We introduce \envname, a novel non-stationary environment to evaluate multi-agent reinforcement learning (MARL) without explicit message passing. This environment features complex agent learning interactions and state dynamics
and is implemented using PettingZoo \cite{terry2021pettingzoo}, a platform that supports creating environments for MARL scenarios.

In \envname, agents are situated in a 32$\times$32 grid-world, populated with three colors of ``blocks'' that yield a \textit{shared} reward to both agents when either of them moves into its grid cell (agents can move up, down, left, and right). At every time-step, one of the three colors gives a $+1$ reward (the ``goal'' block), while the other two yield a $-1$ reward (``incorrect'' blocks). The positive and negative reward values can be customized. Also, the ``goal'' block switches with a default probability of 2.08\% at each time step (see \ref{goal-block-switch-probability} for the computation explanation), introducing non-stationarity but it is slow enough such that the ``follower'' agent can reach it if it switches. Overall, the objective is for the follower to learn to travel to the nearest but different goal block grid square from the leader. Agents can't move through each other, i.e. blocking is possible but we did not observe this phenomenon when visualizing trajectories. Whenever a block is collected in \envname, a new block of the same color is spawned in a random empty cell, maintaining a constant block density and uniform color distribution throughout time. This contributes to \envname's non-stationarity beyond agent movement dynamics. Below we define the set of environment factors we develop that are incorporated in our experiments:

\input{TabsNFigs/tables/env_compare}
\label{subsubsec:environmental}

\paragraph{Customizability} Users of \envname can configure the environment size, the reward and penalty ascribed by the goal and incorrect blocks, the probability of the goal block changing, whether the goal is asymmetric, the density of blocks filling the grid, and the reward shaping functions applied. The latter two affect reward sparsity, which defaults to filling 10\% of the grid with three colors uniformly at random. The  is also customizable. Providing options to toggle non-stationarity and asymmetry make \envname a valuable resource for comparing RL algorithms between multiple levels of environment complexity. 

\paragraph{Asymmetry} In the multi-agent setting of \envname, we define one agent to be the ``leader'' and the other to be the ``follower'' to simulate the scenario where a robot follower must assist a human leader. The leader is always informed which block color is the goal, but this is hidden from the follower when the asymmetric mode is toggled. Adding asymmetry makes this environment significantly more challenging, as the follower must learn to infer the leader's goal from the leader's trajectory only. We also highlight that hidden goals play a crucial role in determining the effectiveness of agents using social learning, i.e. learning through observing the actions of other agents in a shared environment \cite{laland2004social, henrich2003evolution}. Without considering hidden goals, agents may struggle to achieve certain objectives through individual exploration or adapt effectively to new environments. 

\begin{figure}[h]
\begin{center}
\caption{Agent learning in a symmetric vs. asymmetric scenario.}
\includegraphics[width=\linewidth]{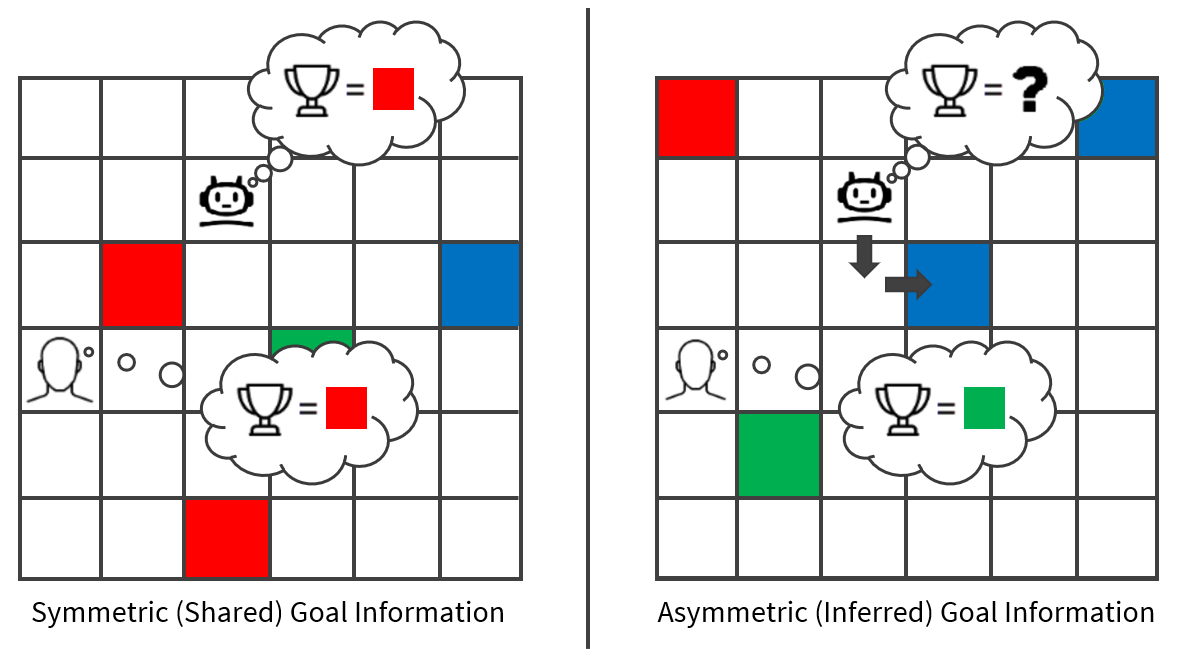}
\end{center}
    \label{fig:asymmetry_vis}
\end{figure}

\paragraph{State Representation} We represent states in a grid representation format, with 5 channels of 32$\times$32 matrices. Each channel is a binary mask for every cell on the 32$\times$32 board, representing whether it contains the color or agent (3 block position and 2 agent position masks result in 5 channels). Additionally, the goal color is provided as a one-hot vector where each index represents a color present in \envname. 

\paragraph{Cost of Exploration} We consider three cases of \envname with an expected reward that is either positive, zero, or negative (in expectation, varying the cost of exploration). See \ref{appendix:reward_values} for the specific reward values.

\paragraph{Penalty Annealing} To lower the cost of exploration at the beginning while the agents are still learning, we linearly raise the reward penalty for collecting blocks of incorrect colors, i.e. the block penalty coefficient increases from $0$ to $1$ between timesteps 4M and 10M, of the 80M total timesteps. 

\paragraph{Distance Reward Shaping}
We introduce a constant penalty when the follower is too close to the leader, incentivizing the follower to collect blocks not targeted by the leader. We use a threshold of 10 Manhattan distance, roughly $\frac{1}{3}$ of the environment length.

\paragraph{Potential Field Reward Shaping}
Another reward shaping term that we develop is inspired by electric potential fields. Just like charged particles, blocks emit a negative or positive value, which increases in magnitude as an agent approaches them. This eliminates sparse rewards, while maintaining an order of magnitude greater reward for collecting a goal block.

\paragraph{Goal Block Switch Probability}
Each environment step, \envname has a chance to switch the color of the goal block, which simulates the leader changing their goal. 
\label{goal-block-switch-probability}

\subsection{Agent Implementation}

We implement our agents as Actor-Critic neural networks \cite{haarnoja2018soft}, with a shared portion that computes features using convolutional and linear layers before feeding into the separate policy and value networks. We use stride size of 1 for all convolutions, Leaky ReLU \cite{maas2013rectifier} as the activation function following all convolution layers, and TanH \cite{LeCun1989GeneralizationAN} following all linear layers. We reference \citet{ndousse2021emergent} for our initial architecture implementation, modifying it to use a stride of 1 for all convolutions (since our state representation is symbolic rather than pixel-based), concatenate a vector to signal the goal block, and add an auxiliary objective network as described in \ref{goal-concatenation}.

\subsubsection{Model Architecture}

\textbf{Shared Convolutional Network:}
\begin{itemize}
    \item \textit{Conv2d Layers}
    \begin{itemize}
        \item Input channels: 5 → 32 → 64
        \item Kernel size: 3 for all layers
    \end{itemize}
\end{itemize}

\noindent \textbf{Shared Projection and Feature Network:}
\begin{itemize}
    \item \textit{Linear Layers} 
    \begin{itemize}
        \item Input: $43264 + 3$ → 192 → 192 → 192
        \item The first layer adds a one-hot vector of goal information (+3 to input dimension). For asymmetric followers, zeros are concatenated instead.
        \item Alternative architecture: Goal information is concatenated at the input of the second layer (see Appendix~\ref{appendix:goal-concatenation}).
    \end{itemize}
    \item \textit{LSTM (if asymmetric)}
    \begin{itemize}
        \item Input: 192 → 192
    \end{itemize}
\end{itemize}

\textbf{Value Network:}
\begin{itemize}
    \item \textit{Linear Layers}
    \begin{itemize}
        \item Input: 192 → 64 → 64 → 1
    \end{itemize}
\end{itemize}

\textbf{Policy Network:}
\begin{itemize}
    \item \textit{Linear Layers}
    \begin{itemize}
        \item Input: 192 → 64 → 64 → 4
    \end{itemize}
\end{itemize}

\textbf{Auxiliary Network (Goal Color Prediction):}
\begin{itemize}
    \item \textit{Linear Layers}
    \begin{itemize}
        \item Input: 192 → 64 → 3
    \end{itemize}
\end{itemize}

% \vspace{0.5em}

\noindent We now discuss two architecture factors: the auxiliary supervised loss and goal color concatenation.

\paragraph{Auxiliary Supervised Loss}
\label{sub3sec:aux_sup_loss}
We add an auxiliary supervised prediction task, in which a two-layer MLP predicts the goal block color from an input of a feature representation output by the shared network. Auxiliary model-based loss objectives have been extensively studied in the RL literature, and in line with previous works we find that this component improves training performance and stability \cite{ndousse2021emergent,ke2019learning,weber2018imaginationaugmented,shelhamer2017loss,jaderberg2016reinforcement,krupnik2019multiagent,hernandezleal2019agent}.
Unless specified otherwise, we use a default coefficient of $\kappa = 0.2$ to add the supervised cross-entropy loss to the main PPO loss. 
With this additional objective, the full loss is described by the below equation, where $c_i = 1$ if the $i$'th color is the current goal, or $0$ otherwise.

\begin{equation}
    \label{eq:supervised_loss}
    L = L_{\text{PPO}} - \kappa \sum_{i = 1}^{3} c_i \cdot \text{log } \hat{c}_i
\end{equation}

\paragraph{Goal Color Concatenation} \label{goal-concatenation} Specifically, in the forward pass, we append an one-hot vector describing the current color of the goal block. This vector is concatenated either ``Early'', immediately after flattening the output from the convolutional network, or ``Late'', after projecting the flattened output into a lower-dimensional space.

\noindent As our goal is to produce agents which can cooperate effectively, the performance metric is the sum of rewards achieved by each agent. We train agents using Independent Proximal-Policy Optimization (IPPO) \cite{schulman2017proximal} to ensure that the individual rewards observed by each agent are correlated with its own actions. For this reason, we don't use Multi-Agent PPO (MAPPO) because the value and policy networks between the leader and follower can't be shared.

%% file: TabsNFigs/tables/env_compare.tex
\begin{table}[t!]
    \centering
    \small

\renewcommand{\arraystretch}{1.5}  % Adjust the value to increase/decrease vertical padding

    \caption{Comparisons between \envname and related environments. ``Non-stationary'' refers to whether the core environment changes over time, without considering the non-stationarity introduced by multiple learning agents. In our case, blocks reappear in the environment and the goal block color may switch. All of these environments have sparse rewards except Multi-Agent Particle.}

\begin{tabular}{|l|p{1.7cm}|p{2.1cm}|p{1.2cm}|p{1.2cm}|}
    \hline
        &                                    \textbf{Observability}  & \textbf{Non-Stationary} &  \textbf{Hidden Goals}   \\ \hline
                            Overcooked             & Full                   & No               &    No                    \\ \hline
                            GoalCycle              & Full                   & No               &    Yes                   \\ \hline
                            Multi-Agent Particle   & Partial/Full           & Yes              &    No                    \\ \hline
                            Level-Based Forage     & Partial/Full           & No               &    No                    \\ \hline
                            Multi-Robot Warehouse  & Partial                & No               &    No                    \\ \hline
                            Coingrid               & Full                   & No               &    No                    \\ \hline
                            \rowcolor[HTML]{47A7F7}         \envname               & Partial/Full           & Yes              &    Yes                   \\ \hline
    \end{tabular}

    \label{tab:compare_envs}
\vspace{-0.5cm}
\end{table}

%% file: sections/04_experiments.tex
\section{Experiments}
\label{sec:experiments}

As this work is exploratory in nature, we focus on completing a breadth of experiments with the resources available rather than running fewer experiments with many seeds. While \envname supports an arbitrary number of leader and follower agents, we focus on the 1:1 scenario emulating the simplest human-agent collaboration case. For all experiments, we use the default block density of 10\% and goal switch probability of 2\% unless otherwise mentioned. We also fix a set of standard hyperparameters for the IPPO algorithm, shown in \ref{appendix:ppo_params}, maintaining a constant learning rate rather than incorporating a schedule. 

\subsection{Baselines}
We use the symmetric setting as a baseline environment to the asymmetric case, which is the main subject of our investigation, and have a baseline leader and follower agent that use the A* search algorithm, where the cost of a path is the length of the shortest path to a goal block (ties are broken arbitrarily) \footnote{This is an admissible heuristic, so this implementation of A* is guaranteed to perform optimally.}. The follower also uses a greedy A* policy, but only updates its current goal with the true reward color when the leader collects a block\footnote{The follower's first actions of an episode are to move to any neighboring empty space until the leader collects a block.}.\footnote{It would be interesting to consider the maximum performance which would be an oracle agent that, in addition to knowing the shortest path to blocks, would have knowledge on where, when, and what color new blocks would spawn, and effectively be able to utilize this information.}

\label{baseline_agents}

\subsection{Findings}
\label{subsec:results}

\begin{figure}[h]
\begin{center}
\includegraphics[width=\linewidth]{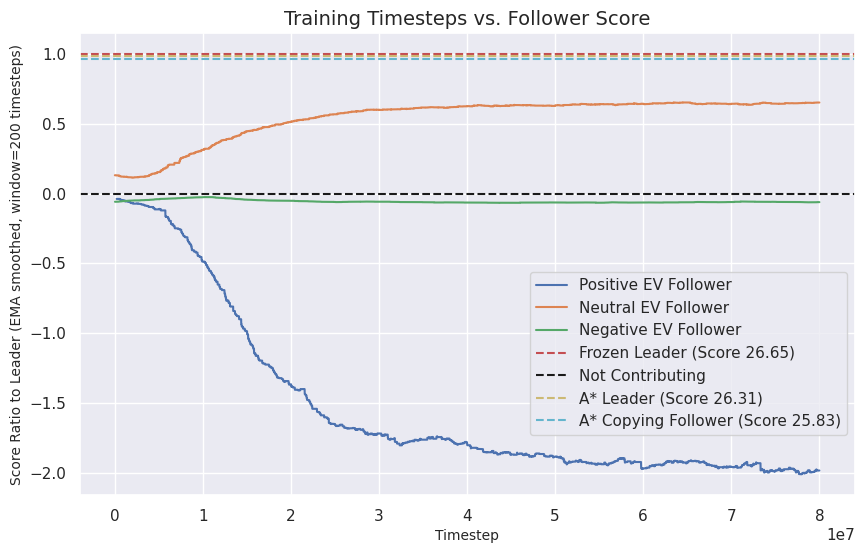}
\end{center}
    \caption{Using a frozen expert leader trained with IPPO, we train a cold-started follower varying the training reward structure such that random block collection would have positive, neutral, and negative expected value (EV). Dotted lines are baselines of the expert leader, A* search leader, and A* search follower which routes to the last color picked up by the A* leader. We use seed 0 for these comparisons, except for the A* agent scores which are averaged across 100 seeds.}
    \label{fig:random_EV_compare}
    \vspace{-0.75em}
\end{figure}

\paragraph{Changing the cost of exploration causes poor agent behavior except in neutral mode.}
\label{cost-of-exploration}

Figure \ref{fig:random_EV_compare} shows the effect of training an IPPO follower in asymmetric environments with a positive, zero, and negative expected cost of exploration, learning with a \textit{frozen} leader policy trained with the best hyperparameters from Table \ref{tab:symmetric_results}. A frozen leader agent operates with a fixed, pre-trained policy and probabilistically switches the goal it seeks rather than with a predictable schedule. When the follower is trained with a fixed leader policy, \envname retains multi-agent dynamics because the follower must adapt to the leader agent's actions. 
 
We train the follower without an auxiliary objective or penalty annealing to purely explore the effect of varying the reward structure. Upon visualizing the environments, we observe the follower agent behavior:
\begin{enumerate}
    \item In an environment with a positive expected value of exploration, the follower learns to pick up all blocks, whether they are the color of the goal or not.
    \item In an environment with a negative expected value of exploration, the follower learns to avoid picking up any blocks, and does not distinguish which are the goal.
    \item Only in an environment with an expected cost of exploration of zero does the follower learn to pick up the correct block color often, and less frequently collecting incorrect blocks. 
\end{enumerate}

Note that in these experiments, the goal block color switches randomly, implying that the follower in the neutral environment occasionally learned to correctly observe the leader's behavior to inform which goal to collect. We hypothesize the pessimistic follower avoids collecting any blocks due to a sparsity of reward: it is rare that the leader collects the goal block, and the follower randomly chooses the goal block shortly after. To maximize reward in an environment where uninformed block pickup occurs, the follower avoids picking up blocks and incurring the penalty. On the contrary, the optimistic follower collects every block in its path. We infer the follower learns a local optimum that collecting blocks uniformly at random yields positive reward, and thus never learns the global optimum of only collecting the goal block.

Even in a neutral environment, which we find to be the best environment for learning a follower agent using IPPO, the agent still collects incorrect blocks. In a real-world scenario with high consequences for mistaken actions, this would be unacceptable for a human helper. Though the Negative EV follower never collects incorrect blocks, it also never collects goal blocks, which is harmless but not helpful. We believe the real world is a pessimistic (Negative EV) environment if pursuing random goals, where many actions exist that cause harm, and only a few are helpful. This, combined with IPPO learning poorly in a pessimistic environment, motivates our study of how to get a follower to learn in a pessimistic environment for the rest of the paper. We observe that the frozen leader score is approximately on-par with the baseline A* leader score, showing that IPPO learns a policy close to classical heuristic-based search algorithms but is capable of learning more as it's inherently "blind" like a greedy algorithm. The A* copying follower performance, on the other hand, significantly outperforms all follower agents trained with IPPO, demonstrating that even with an expected cost of exploration of zero, IPPO is insufficient to learn to infer the leader's goal.

% Though we have successfully demonstrated that a follower can help the leader by, on average, improving total reward if trained in a neutral environment, we do not believe this is representative of the real world. In the real world, a human also has a specific goal in mind, but the expected value of randomly acting is negative, \textit{not} neutral. We therefore study how to improve SOTA, IPPO, which failed in the Negative EV environment, for the rest of the paper.

% MAKE A STRONGER CLAIM: neutral does not fully solve, real world consequences would be severe even on neutral

% is unsurprising since we don't expect there to be many cases described in \autoref{baseline_agents}. Compared to the other follower scores,
% this plot also demonstrates that despite all the reward hacking IPPO is insufficient to learn well in \envname.

\begin{table}[h!]
  \centering
  \small

  \caption{``Converged Sum Reward'' is the final net reward over 128 timesteps averaged over 16 environments. All runs use 10\% block density, non-stationary goal block color with 2\% chance of switching at each step, and \emph{symmetric} information between the leader and follower. We take the converged reward values after 80M train timesteps and average across 3 seeds.}

  \begin{tabular}{l@{\hspace{0.95\tabcolsep}}
                  l@{\hspace{0.95\tabcolsep}}
                  |
                  r}
    \textbf{Auxiliary Loss } $\kappa$ & \textbf{Penalty Annealing} & \textbf{Converged $\Sigma$ Reward} \\
    0.2 & 4M-10M & \textbf{48.8} \\
    0.2 & None & -0.7 \\
    0 & None & -1.0 \\
    0 & 4M-10M & 32.8 \\
  \end{tabular}
  
  \label{tab:symmetric_results}
\end{table}

\paragraph{Annealing the incorrect block penalty enables learning through early exploration.}

\label{annealing}
Early exploration is key for the agents to learn. We find that training with a constant incorrect penalty results in a negative reward after 80M timesteps (\ref{tab:symmetric_results}). Without penalty annealing, because the environment is pessimistic with negative expected value, agents converge adversely toward avoiding blocks altogether whereas introducing penalty annealing creates a better learning dynamic.

\paragraph{Supervised goal prediction stabilizes training with sparse rewards.} 

We find that there is a clear benefit of the auxiliary goal prediction objective through an ablation with 5\% block density, the sparsest environment configuration that we explore. In this experiment we use symmetric information, providing the current goal information to both the leader and follower agents. With such sparse reward signal, naive PPO training collapses to inaction once the block penalty starts increasing at 4M timesteps. Similar to the case when training without penalty annealing, these agents degenerate to a behavior of avoiding collecting any blocks at all, minimizing the penalties incurred but failing to achieve any positive reward.

However, when including the supervised auxiliary objective with $\kappa = 0.2$, the agents are able to recover and learn to collect the current goal blocks while avoiding the other colors. Intuitively, the supervised loss forces the feature representation output by the shared network to encode the goal information, thereby allowing the policy network to perform optimally given the current goal. We additionally conduct an experiment with a higher coefficient $\kappa = 0.4$ but
find that weighting the auxiliary loss too heavily reduces performance, demonstrating the importance of tuning this hyperparameter.

\paragraph{Unbalanced learning collapses to a single-agent solution.}

We observe that when one agent learns faster than the other, it collects goal blocks which are close to the other agent, removing opportunities to receive positive reward signal, resulting in a single-agent solution where the slower agent does not collect any blocks. As an attempt to get agents to learn in the asymmetric setting, we introduced distance reward shaping, penalizing the leader from being close to the follower. While this does allow the follower to explore during the portion of training with no block penalty, once the penalty starts increasing it again collapses to not collecting any blocks.

\paragraph{Warmstarting with full information is promising for transferring to asymmetric follower.} 

One intuitive result is that in the asymmetric setting with distance reward shaping, when warmstarting a follower from a follower and leader with full goal information for 80M timesteps and $\kappa=0.2$, the agents score better than a follower that's freshly initialized. All parameter configurations (\ref{appendix:warm-start}) had similar results.

\paragraph{Varying goal switch probabilities shows \envname complexity independent of goal changing.}

We perform an ablation of goal switch probability, comparing $P = 0.00, 0.25, 0.50, 0.75$. Note that $P=0.00$ should be the best performing case, when the leader's goal never switches in this setting, we expect the follower to learn to always pick up a fixed color. In reality, we see that the difference in reward across the different block switch probabilities does not significantly vary. Moreover, we hypothesize that the follower might be able to learn approximately when the goal color will switch. Overall, this suggests that \envname is challenging for IPPO not only due to switching the goal, but also due to other factors like sparse rewards.

\paragraph{Dense reward shaping does not solve goal inference for pessimistic environments.}

To combat reward sparsity, we experimented with applying a reward shaping term which provides rewards based on proximity to goal blocks and non-goal blocks, similar to an electric potential field where goal blocks represent positive potential and incorrect blocks represent negative potential. Still, we see that the follower agent performance is poor; we hypothesize this to be the case because despite denser rewards, the negative expected cost of exploration still heavily punishes exploration.

%% file: sections/05_discussion.tex
\section{Conclusion}

We present \envname, a novel multi-agent environment with changing and hidden goals. We find that it is a challenging MARL environment that SOTA IPPO cannot solve out-of-the-box, and analyze several architectural, environmental, and algorithmic factors that make the environment easier or harder for agents. Notably, we see that the cost of exploration directly affects an agent's ability to learn while collaborating. By making a simplified gridworld environment to focus on studying the keys aspects of this problem from a cooperation perspective and while abstracting away details like image understanding or low-level motor control, \envname is another step towards human-AI cooperation and social learning. We hope this research helps spur other works to gain more insights into how costlier individual exploration raises the incentive to rely on social learning and enables the development of AI agents that can adaptively help humans without explicit feedback in real-time and realistic settings where goals are constantly changing. This can impact a wide range of fields and industries including robotic surgery and household assistance. Specifically, a robotic surgery assistant can better respond to patient-specific comfort levels, which could lead to improved patient outcomes, or a robot could help another human set the table while not getting in the way.

\subsection{Limitations and Future Work}
% WE TRIED (on negative env)
% * block swap probability
% * reward shaping with proximity
% * reward shaping with potential field
% * Reward Signal Distribution
%     > hemispheres, follower still cannot learn in negative EV
% * leader AND follower annealing from a warm started leader checkpoint
% * symmetric vs asymmetric, goal sharing for follower and leader vs not
% * leader and follower learning from zero
% \todo{remove, enumerate if needed, more for us while writing}

Apart from running more experiments with different seeds and determining optimal hyperparameter configurations, avenues for future work broadly fall into two categories: environmental and algorithmic design. These support developing multi-agent systems that collaborate more effectively with humans and other agents in zero-shot settings.

\subsubsection{Environmental Developments}

There are various environment developments that could further evaluate agents' capabilities to adapt more robust strategies by introducing uncertainty. To examine the effects of goal rarity, non-uniform block distributions and stochastic rewards tied to block color sampled from distributions with varying means and variances, could be implemented. An alternative approach to challenge agents could involve structured mazes or hemispherical designs where blocks collected in one hemisphere respawn in the other. Adopting curriculum-based approaches, like POET \cite{poet}, could facilitate progressive learning by generating adversarial environments that push agents' abilities. Lastly, real-world scenarios often involve humans outnumbering autonomous agents, requiring followers to prioritize which leaders to assist. Future studies could explore settings where followers assist multiple leaders, each with different goals, or scenarios where a follower faces conflicting objectives from two leaders. These niche but critical challenges are essential for developing robust multi-agent coordination.

\subsubsection{Algorithmic Developments}

Advancing algorithm developments beyond IPPO is critical to address the challenges presented by \envname's complex multi-agent scenarios. One potential improvement is warm-starting the follower agent by behavior cloning the A* copying baseline. Another method that can likely surpass behavior cloning is leveraging diverse policy data collected from baseline A* agents for Implicit Q-Learning. An extension of IQL specifically is using a COMA-inspired objective \cite{foerster2017counterfactual}, such as $Q(a_R, s) = Q(a_H, a_R, s) - Q(a_H, s)$, which could help the follower isolate its unique contribution to the reward. Online inverse reinforcement learning (IRL) represents another promising direction for goal inference. Techniques like BASIS \cite{abdulhai2022basis} could enable the follower to infer the leader's current goal explicitly by learning a model of the leader’s reward function. This approach could complement or outperform auxiliary supervised objectives by embedding goal inference directly into the agent’s policy. 

%% file: sections/06_acknowledgements.tex
\section{Acknowledgements}

We would like to thank Professor Natasha Jaques for guidance on this project and Sriyash Poddar for the insightful discussions.

%% file: sections/appendix.tex
\section{Appendix}
\label{appendix}

\subsection{Cost of Exploration Reward Values}
Below are the reward values for the three cases of \envname that are differentiated by having a positive (optimistic), zero (neutral), or negative (pessimistic) expected value for collecting a random block, respectively.
\label{appendix:reward_values}
\begin{table}[h!]
  \small

  \begin{tabular}{l@{\hspace{0.95\tabcolsep}}
                  l@{\hspace{0.95\tabcolsep}}
                  r}

    \textbf{Goal Reward} & \textbf{Incorrect Reward} \\
    +4 & -1 \\
    +2 & -1 \\
    +1 & -1
  \end{tabular}
\end{table}

% \vspace{-1em}

\subsection{PPO Parameters}
\label{appendix:ppo_params}
% \hspace{0pt}
\begin{table}[h!] % Use [h!] to force placement "here" immediately
\begin{tabular}{l c}
    \hline
    Parameter         & Value \\
    \hline
    LR                & 1e-4 \\
    N. Envs           & 16 \\
    N. Rollout Steps  & 128 \\
    Gamma             & 0.99 \\
    GAE Lambda        & 0.95 \\
    N. Minibatches    & 4 \\
    Update Epochs     & 4 \\
    Clip Param        & 0.2 \\
    Entropy Coef      & 0.01 \\
    Value Coef        & 0.5 \\
    Target KL         & 0.01 \\
    \hline
\end{tabular}
\end{table}

% \vspace{-1em}

\subsection{Goal Block Switch Probability}
The default block switching probability parameter is $\frac{2}{3} \cdot \frac{1}{32} \approx 2.08\%$. The $\frac{2}{3}$ is motivated by the $\frac{1}{3}$ chance to randomly select the same color, while the $\frac{1}{32}$ gives the leader enough time steps to collect a new goal block color before switching again (in expectation, it takes 32 environment steps before the goal block switches, and 32 steps is enough to cross half the default 32x32 environment). 

\subsection{Warmstarting Experiment Parameter Details}
\label{appendix:warm-start}
For warmstarting, we apply the distance penalty reward shaping term for a distance threshold of 10, penalty 0.25, for 20M timesteps, and penalty annealing from 10M to 20M timestep. We also try penalty 0.5 for distance threshold 10 for 40M time steps and penalty annealing for 4M to 10M.

\subsection{Additional Ablation: Early vs. Late Goal Concatenation}
\paragraph{Concatenating goal information earlier in the forward pass is insignificant.}
\label{appendix:goal-concatenation}

We find that, in an environment with 10\% block density, concatenating the one-hot goal information vector immediately after flattening the output from the convolutional network doesn't have a substantially higher shared reward than concatenating the vector after projecting the flattened output into a lower-dimensional space. We hypothesize that the difference is negligible because either way, there are enough layers after appending to appropriately encode the goal information into the feature representation.

\subsection{Hardware Details}
\envname takes approximately 2 seconds per 1,000 time-steps on a A40 or L40 GPU. With some additional code improvements, this runtime can be optimized much further.

%% file: main.bbl
\begin{thebibliography}{36}
\providecommand{\natexlab}[1]{#1}
\providecommand{\url}[1]{\texttt{#1}}
\expandafter\ifx\csname urlstyle\endcsname\relax
  \providecommand{\doi}[1]{doi: #1}\else
  \providecommand{\doi}{doi: \begingroup \urlstyle{rm}\Url}\fi

\bibitem[Abdulhai et~al.(2022)Abdulhai, Jaques, and Levine]{abdulhai2022basis}
Marwa Abdulhai, Natasha Jaques, and Sergey Levine.
\newblock Basis for intentions: Efficient inverse reinforcement learning using past experience, 2022.

\bibitem[Albrecht and Ramamoorthy(2015)]{albrecht2015gametheoretic}
Stefano~V. Albrecht and Subramanian Ramamoorthy.
\newblock A game-theoretic model and best-response learning method for ad hoc coordination in multiagent systems, 2015.

\bibitem[Carroll et~al.(2019)Carroll, Shah, Ho, Griffiths, Seshia, Abbeel, and Dragan]{overcooked}
Micah Carroll, Rohin Shah, Mark~K. Ho, Thomas~L. Griffiths, Sanjit~A. Seshia, Pieter Abbeel, and Anca~D. Dragan.
\newblock On the utility of learning about humans for human-ai coordination.
\newblock \emph{CoRR}, abs/1910.05789, 2019.
\newblock URL \url{http://arxiv.org/abs/1910.05789}.

\bibitem[de~Witt et~al.(2020)de~Witt, Gupta, Makoviichuk, Makoviychuk, Torr, Sun, and Whiteson]{dewitt2020independentlearningneedstarcraft}
Christian~Schroeder de~Witt, Tarun Gupta, Denys Makoviichuk, Viktor Makoviychuk, Philip H.~S. Torr, Mingfei Sun, and Shimon Whiteson.
\newblock Is independent learning all you need in the starcraft multi-agent challenge?, 2020.
\newblock URL \url{https://arxiv.org/abs/2011.09533}.

\bibitem[Dennis et~al.(2021)Dennis, Jaques, Vinitsky, Bayen, Russell, Critch, and Levine]{dennis2021emergent}
Michael Dennis, Natasha Jaques, Eugene Vinitsky, Alexandre Bayen, Stuart Russell, Andrew Critch, and Sergey Levine.
\newblock Emergent complexity and zero-shot transfer via unsupervised environment design, 2021.

\bibitem[Foerster et~al.(2017)Foerster, Farquhar, Afouras, Nardelli, and Whiteson]{foerster2017counterfactual}
Jakob Foerster, Gregory Farquhar, Triantafyllos Afouras, Nantas Nardelli, and Shimon Whiteson.
\newblock Counterfactual multi-agent policy gradients, 2017.

\bibitem[Haarnoja et~al.(2018)Haarnoja, Zhou, Abbeel, and Levine]{haarnoja2018soft}
Tuomas Haarnoja, Aurick Zhou, Pieter Abbeel, and Sergey Levine.
\newblock Soft actor-critic: Off-policy maximum entropy deep reinforcement learning with a stochastic actor, 2018.

\bibitem[Henrich and McElreath(2003)]{henrich2003evolution}
Joseph Henrich and Richard McElreath.
\newblock The evolution of cultural evolution.
\newblock \emph{Evolutionary Anthropology: Issues, News, and Reviews: Issues, News, and Reviews}, 12\penalty0 (3):\penalty0 123--135, 2003.

\bibitem[Hernandez-Leal et~al.(2019)Hernandez-Leal, Kartal, and Taylor]{hernandezleal2019agent}
Pablo Hernandez-Leal, Bilal Kartal, and Matthew~E. Taylor.
\newblock Agent modeling as auxiliary task for deep reinforcement learning, 2019.

\bibitem[Jaderberg et~al.(2016)Jaderberg, Mnih, Czarnecki, Schaul, Leibo, Silver, and Kavukcuoglu]{jaderberg2016reinforcement}
Max Jaderberg, Volodymyr Mnih, Wojciech~Marian Czarnecki, Tom Schaul, Joel~Z Leibo, David Silver, and Koray Kavukcuoglu.
\newblock Reinforcement learning with unsupervised auxiliary tasks, 2016.

\bibitem[Jaderberg et~al.(2018)Jaderberg, Czarnecki, Dunning, Marris, Lever, Casta{\~n}eda, Beattie, Rabinowitz, Morcos, Ruderman, Sonnerat, Green, Deason, Leibo, Silver, Hassabis, Kavukcuoglu, and Graepel]{Jaderberg2018HumanlevelPI}
Max Jaderberg, Wojciech~M. Czarnecki, Iain Dunning, Luke Marris, Guy Lever, Antonio~Garc{\'i}a Casta{\~n}eda, Charlie Beattie, Neil~C. Rabinowitz, Ari~S. Morcos, Avraham Ruderman, Nicolas Sonnerat, Tim Green, Louise Deason, Joel~Z. Leibo, David Silver, Demis Hassabis, Koray Kavukcuoglu, and Thore Graepel.
\newblock Human-level performance in 3d multiplayer games with population-based reinforcement learning.
\newblock \emph{Science}, 364:\penalty0 859 -- 865, 2018.
\newblock URL \url{https://api.semanticscholar.org/CorpusID:49561741}.

\bibitem[Jain and Argall(2018)]{recursivebayesian}
Siddarth Jain and Brenna Argall.
\newblock Recursive bayesian human intent recognition in shared-control robotics.
\newblock In \emph{2018 IEEE/RSJ International Conference on Intelligent Robots and Systems (IROS)}, pages 3905--3912, 2018.
\newblock \doi{10.1109/IROS.2018.8593766}.

\bibitem[Jeon et~al.(2020)Jeon, Losey, and Sadigh]{jeon2020shared}
Hong~Jun Jeon, Dylan~P. Losey, and Dorsa Sadigh.
\newblock Shared autonomy with learned latent actions, 2020.

\bibitem[Ke et~al.(2019)Ke, Singh, Touati, Goyal, Bengio, Parikh, and Batra]{ke2019learning}
Nan~Rosemary Ke, Amanpreet Singh, Ahmed Touati, Anirudh Goyal, Yoshua Bengio, Devi Parikh, and Dhruv Batra.
\newblock Learning dynamics model in reinforcement learning by incorporating the long term future, 2019.

\bibitem[Krupnik et~al.(2019)Krupnik, Mordatch, and Tamar]{krupnik2019multiagent}
Orr Krupnik, Igor Mordatch, and Aviv Tamar.
\newblock Multi-agent reinforcement learning with multi-step generative models, 2019.

\bibitem[Laland(2004)]{laland2004social}
Kevin~N Laland.
\newblock Social learning strategies.
\newblock \emph{Animal Learning \& Behavior}, 32\penalty0 (1):\penalty0 4--14, 2004.

\bibitem[LeCun(1989)]{LeCun1989GeneralizationAN}
Yann LeCun.
\newblock Generalization and network design strategies.
\newblock 1989.
\newblock URL \url{https://api.semanticscholar.org/CorpusID:59861896}.

\bibitem[Maas et~al.(2013)Maas, Hannun, Ng, et~al.]{maas2013rectifier}
Andrew~L Maas, Awni~Y Hannun, Andrew~Y Ng, et~al.
\newblock Rectifier nonlinearities improve neural network acoustic models.
\newblock In \emph{Proc. icml}, volume~30, page~3. Atlanta, GA, 2013.

\bibitem[Mordatch and Abbeel(2018)]{mordatch2018emergence}
Igor Mordatch and Pieter Abbeel.
\newblock Emergence of grounded compositional language in multi-agent populations, 2018.

\bibitem[Ndousse et~al.(2021)Ndousse, Eck, Levine, and Jaques]{ndousse2021emergent}
Kamal Ndousse, Douglas Eck, Sergey Levine, and Natasha Jaques.
\newblock Emergent social learning via multi-agent reinforcement learning, 2021.

\bibitem[Nikolaidis et~al.(2017)Nikolaidis, Hsu, and Srinivasa]{Nikolaidis2017HumanrobotMA}
Stefanos Nikolaidis, David Hsu, and Siddhartha~S. Srinivasa.
\newblock Human-robot mutual adaptation in collaborative tasks: Models and experiments.
\newblock \emph{The International Journal of Robotics Research}, 36:\penalty0 618 -- 634, 2017.
\newblock URL \url{https://api.semanticscholar.org/CorpusID:7274323}.

\bibitem[Papoudakis et~al.(2021)Papoudakis, Christianos, Schäfer, and Albrecht]{epymarl}
Georgios Papoudakis, Filippos Christianos, Lukas Schäfer, and Stefano~V. Albrecht.
\newblock Benchmarking multi-agent deep reinforcement learning algorithms in cooperative tasks.
\newblock In \emph{Proceedings of the Neural Information Processing Systems Track on Datasets and Benchmarks (NeurIPS)}, 2021.
\newblock URL \url{http://arxiv.org/abs/2006.07869}.

\bibitem[Puig et~al.(2021)Puig, Shu, Li, Wang, Liao, Tenenbaum, Fidler, and Torralba]{puig2021watchandhelp}
Xavier Puig, Tianmin Shu, Shuang Li, Zilin Wang, Yuan-Hong Liao, Joshua~B. Tenenbaum, Sanja Fidler, and Antonio Torralba.
\newblock Watch-and-help: A challenge for social perception and human-ai collaboration, 2021.

\bibitem[Raileanu et~al.(2018)Raileanu, Denton, Szlam, and Fergus]{raileanu2018modeling}
Roberta Raileanu, Emily Denton, Arthur Szlam, and Rob Fergus.
\newblock Modeling others using oneself in multi-agent reinforcement learning, 2018.

\bibitem[Schulman et~al.(2017)Schulman, Wolski, Dhariwal, Radford, and Klimov]{schulman2017proximal}
John Schulman, Filip Wolski, Prafulla Dhariwal, Alec Radford, and Oleg Klimov.
\newblock Proximal policy optimization algorithms, 2017.

\bibitem[Shelhamer et~al.(2017)Shelhamer, Mahmoudieh, Argus, and Darrell]{shelhamer2017loss}
Evan Shelhamer, Parsa Mahmoudieh, Max Argus, and Trevor Darrell.
\newblock Loss is its own reward: Self-supervision for reinforcement learning, 2017.

\bibitem[Shi et~al.(2024)Shi, Hu, Zhao, Sharma, Pertsch, Luo, Levine, and Finn]{shi2024yell}
Lucy~Xiaoyang Shi, Zheyuan Hu, Tony~Z. Zhao, Archit Sharma, Karl Pertsch, Jianlan Luo, Sergey Levine, and Chelsea Finn.
\newblock Yell at your robot: Improving on-the-fly from language corrections, 2024.

\bibitem[Shrestha and Moore(2014)]{Shrestha_2014}
Munik Shrestha and Cristopher Moore.
\newblock Message-passing approach for threshold models of behavior in networks.
\newblock \emph{Physical Review E}, 89\penalty0 (2), February 2014.
\newblock ISSN 1550-2376.
\newblock \doi{10.1103/physreve.89.022805}.
\newblock URL \url{http://dx.doi.org/10.1103/PhysRevE.89.022805}.

\bibitem[Silver et~al.(2017)Silver, Hubert, Schrittwieser, Antonoglou, Lai, Guez, Lanctot, Sifre, Kumaran, Graepel, Lillicrap, Simonyan, and Hassabis]{silver2017mastering}
David Silver, Thomas Hubert, Julian Schrittwieser, Ioannis Antonoglou, Matthew Lai, Arthur Guez, Marc Lanctot, Laurent Sifre, Dharshan Kumaran, Thore Graepel, Timothy Lillicrap, Karen Simonyan, and Demis Hassabis.
\newblock Mastering chess and shogi by self-play with a general reinforcement learning algorithm, 2017.

\bibitem[Smallwood and Sondik(1973)]{Smallwood1973TheOC}
Richard~D. Smallwood and Edward~J. Sondik.
\newblock The optimal control of partially observable markov processes over a finite horizon.
\newblock \emph{Oper. Res.}, 21:\penalty0 1071--1088, 1973.
\newblock URL \url{https://api.semanticscholar.org/CorpusID:43604344}.

\bibitem[Strouse et~al.(2022)Strouse, McKee, Botvinick, Hughes, and Everett]{strouse2022collaborating}
DJ~Strouse, Kevin~R. McKee, Matt Botvinick, Edward Hughes, and Richard Everett.
\newblock Collaborating with humans without human data, 2022.

\bibitem[Terry et~al.(2021)Terry, Black, Grammel, Jayakumar, Hari, Sullivan, Santos, Dieffendahl, Horsch, Perez-Vicente, et~al.]{terry2021pettingzoo}
J~Terry, Benjamin Black, Nathaniel Grammel, Mario Jayakumar, Ananth Hari, Ryan Sullivan, Luis~S Santos, Clemens Dieffendahl, Caroline Horsch, Rodrigo Perez-Vicente, et~al.
\newblock Pettingzoo: Gym for multi-agent reinforcement learning.
\newblock \emph{Advances in Neural Information Processing Systems}, 34:\penalty0 15032--15043, 2021.

\bibitem[Wang et~al.(2019)Wang, Lehman, Clune, and Stanley]{poet}
Rui Wang, Joel Lehman, Jeff Clune, and Kenneth~O. Stanley.
\newblock Paired open-ended trailblazer {(POET):} endlessly generating increasingly complex and diverse learning environments and their solutions.
\newblock \emph{CoRR}, abs/1901.01753, 2019.
\newblock URL \url{http://arxiv.org/abs/1901.01753}.

\bibitem[Weber et~al.(2018)Weber, Racanière, Reichert, Buesing, Guez, Rezende, Badia, Vinyals, Heess, Li, Pascanu, Battaglia, Hassabis, Silver, and Wierstra]{weber2018imaginationaugmented}
Théophane Weber, Sébastien Racanière, David~P. Reichert, Lars Buesing, Arthur Guez, Danilo~Jimenez Rezende, Adria~Puigdomènech Badia, Oriol Vinyals, Nicolas Heess, Yujia Li, Razvan Pascanu, Peter Battaglia, Demis Hassabis, David Silver, and Daan Wierstra.
\newblock Imagination-augmented agents for deep reinforcement learning, 2018.

\bibitem[Weil et~al.(2024)Weil, Bao, Abboud, and Meuser]{weil2024generalizability}
Jannis Weil, Zhenghua Bao, Osama Abboud, and Tobias Meuser.
\newblock Towards generalizability of multi-agent reinforcement learning in graphs with recurrent message passing, 2024.

\bibitem[Yu et~al.(2021)Yu, Velu, Vinitsky, Gao, Wang, Bayen, and Wu]{yu2021surprising}
Chao Yu, Akash Velu, Eugene Vinitsky, Jiaxuan Gao, Yu~Wang, Alexandre Bayen, and Yi~Wu.
\newblock The surprising effectiveness of ppo in cooperative, multi-agent games, 2021.

\end{thebibliography}
